\pdfoutput=1

\documentclass[11pt]{article}

\usepackage[preprint]{acl}
\usepackage{times}
\usepackage{latexsym}
\usepackage{float}

\usepackage[T1]{fontenc}

\usepackage[utf8]{inputenc}

\usepackage{microtype}

\usepackage{inconsolata}
\usepackage{algorithm}
\usepackage{algpseudocode}
\usepackage{relsize} 
\usepackage{bbm}

\usepackage{booktabs}
\usepackage{algorithm}
\usepackage{algpseudocode}
\usepackage{graphicx}
\usepackage{longtable}
\usepackage{tcolorbox}
\usepackage{stackengine}
\usepackage{multirow}
\usepackage{graphicx}
\usepackage{wrapfig}
\usepackage{hyperref}
\usepackage{makecell}
\usepackage{booktabs} 
\usepackage{array}    
\usepackage{geometry} 

\usepackage{tcolorbox}
\tcbuselibrary{breakable}

\usepackage{url}            
\usepackage{amsfonts}       
\usepackage{nicefrac}       
\usepackage{microtype}      
\usepackage{xcolor,colortbl}         
\usepackage{times}
\usepackage{latexsym}
\usepackage{multicol}
\usepackage{blindtext}
\usepackage{tabu}
\usepackage{amsmath, bm}

\usepackage{subcaption}
\usepackage{caption}
\usepackage[normalem]{ulem}
\usepackage{soul}
\usepackage[shortlabels]{enumitem}
\usepackage{array}
\usepackage{pgffor}
\usepackage{textcomp}
\usepackage{amssymb}
\usepackage{pifont}
\usepackage{booktabs}
\usepackage{tabularx}
\usepackage{lipsum}

\definecolor{lightgreen}{rgb}{0.8, 0.95, 0.8}
\definecolor{lightred}{rgb}{0.95, 0.8, 0.8}
\definecolor{naplesyellow}{rgb}{0.98, 0.85, 0.37}
\definecolor{pastelyellow}{rgb}{0.99, 0.99, 0.59}

\newcommand{\method}{\textsc{ThinkTuning}\xspace}

\algrenewcommand\algorithmicindent{0.85em}
\title{\method: Instilling Cognitive Reflections without Distillation}

\author{Aswin RRV \quad Jacob Dineen \quad Divij Handa \quad Md Nayem Uddin\\ \textbf{Mihir Parmar}\quad \textbf{Chitta Baral}\quad \textbf{Ben Zhou} \\ 
Arizona State University\\
\small{\texttt{\{aravik13, chitta, xzhou202\}@asu.edu}}
}

\begin{document}
\maketitle
\begin{abstract}

Recent advances in test-time scaling have led to the emergence of thinking LLMs that exhibit self-reflective behaviors and multi-step reasoning. While RL drives this self-improvement paradigm, a recent study \cite{cognitive_behaviors} shows that RL alone does not truly instill these new reasoning abilities - it merely draws out behaviors already present in the base models. This raises a question: \textit{How can we train the models that don't exhibit such thinking behavior to develop it in the first place?} To this end, we propose \method, a GRPO-based interactive training approach where we augment the rollouts of a student model with the guidance from a teacher model. A simple idea from classroom practice inspires our method: a teacher poses a problem, lets the student try an answer, then gives corrective feedback--enough to point the mind in the right direction and then show the solution. Each piece of feedback reshapes the student's thoughts, leading them to arrive at the correct solution. Similarly, we find that this type of implicit supervision through feedback from a teacher model of the same size improves the reasoning capabilities of the student model. In particular, on average, our method shows a $3.85\%$ improvement over zero-shot baselines across benchmarks, and on MATH-500, AIME and GPQA-Diamond it shows $2.08\%$, $2.23\%$ and $3.99\%$ improvements over the vanilla-GRPO baseline\footnote{Source code is available at \url{https://github.com/3rdAT/ThinkTuning}}.

\end{abstract}

\section{Introduction}
\label{sec:intro}

\begin{figure}[t]
  \centering
  \includegraphics[width=\linewidth]{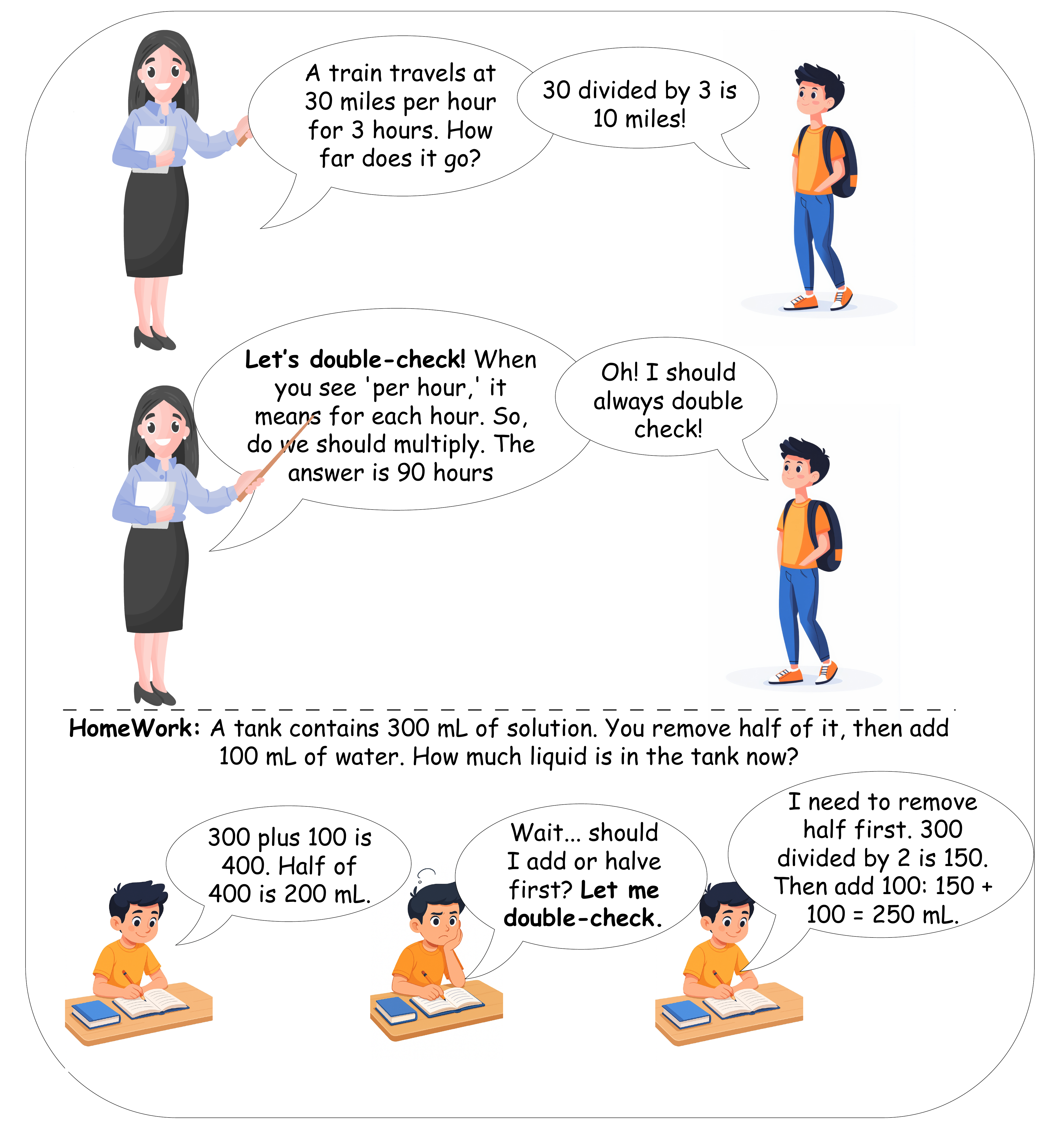}
  \caption{Illustration of \method motivation. 
  Top: teacher poses a math problem, student answers incorrectly, and the teacher offers a short corrective feedback. Bottom: For a new problem, the student recalls the feedback ("double-check") and ends up producing the correct answer.}
  \label{fig:teaser}
\end{figure}

Recent progress in AI research has been driven by advances in scaling the models' parameter count \citep{scaling-laws}. More recently, scaling along the inference-time axis has produced significant performance gains in various complex reasoning tasks \cite{snell2025scaling}. Thinking models such as OpenAI-o-series \citep{openai_o1}, DeepSeek-R1 \citep{guo2025deepseek} and Gemini-Thinking \citep{team2023gemini} are a testament to this, capable of producing long reasoning chains, with sophisticated behaviors like self-reflection, self-correction, and multi-step reasoning. These significant performance gains are attributed to the success of Reinforcement Learning (RL) through simple rule-based rewards. However, online on-policy RL settings face a constraint: sophisticated reasoning behaviors will not emerge unless they are explicitly sampled during training. For example, models like Qwen \citep{yang2025qwen3} often come with strong priors, allowing them to naturally generate sophisticated reasoning behaviors, which RL then amplifies. In contrast, when models lack strong priors, on-policy RL struggles to elicit them. Indeed, a recent study shows that RL applied on the Llama 3.2–family \citep{Llama3-herd} of models struggles to elicit sophisticated reasoning behaviors \citep{cognitive_behaviors}.

In academic settings, cognitive modeling provides a structured approach for shaping both overt (external) and covert (internal-cognitive) behaviors of students through guided interventions by a teacher, typically using verbal mediation \cite{camp1978think}. As illustrated in Fig. \ref{fig:teaser}, imagine a teacher asking, “A train travels at 30 miles per hour for 3 hours. How far does it go?” A hasty student might reply, “30 divided by 3 is 10 miles!” A good teacher not only explains why the answer is incorrect but also imparts a generalizable skill. In this case, the teacher could encourage the student to double-check what “per hour” means and to think carefully about whether they should multiply or divide in similar problems. Particularly, in STEM education literature \cite{chouvalova2024undergraduates}, it has been established that the use of corrective feedback with errorful learning is more beneficial in student learning\footnote{A Quote from \cite{chouvalova2024undergraduates}: 
``{\em Compared to passively reading materials, errorful learning paired with corrective feedback is more beneficial to student learning and retention  \cite{mera2022unraveling,overman2021enhanced}.}}.  Interestingly, recent thinking models often exhibit such behavior of re-checking and self-refining, which makes them better at various reasoning tasks. Presumably, these thinking behaviors emerge in those models solely through RL, as suitable priors are present to help in exhibiting such behavior \cite{cognitive_behaviors}. However, this brings up an important question: \textit{How can we enable models to acquire these types of thinking skills in the absence of suitable priors? And is RL alone sufficient for this task?} 

Drawing inspiration from the example discussed above, we propose \method, a training approach where an active student model learns to think by interacting with a teacher model. Rather than assuming thinking behaviors will emerge during RL, we engineer the training process to induce them. This aligns with how cognitive modeling in educational settings elicits complex reasoning strategies, such as self-reflection, self-correction, and problem-solving among students. 

\method consists of two stages. First, we start by creating a set of few-shot exemplars, each demonstrating an opinion on a student’s response, a reason for that opinion, and a phrase that typically showcases specific cognitive behaviors by solving a problem. Our exemplars capture the most common human self-reflective behaviors: Self-Conflict, Self-Agreement, Self-Critique, and Self-Consultancy. While many other cognitive behaviors exist, we focus on these four because they are well defined \citep{hermans2023dialogical, hermans2011handbook}. Second, we train the student model in an online RL setting with Group Relative Policy Optimization (GRPO) \citep{shao2024deepseekmath}. At each iteration, the student model generates $n$ rollouts, from which a subset of $\gamma$ rollouts is randomly selected. These selected rollouts are passed to the few-shot teacher model to obtain feedback and phrases showcasing the cognitive skill. The feedback is then appended to the corresponding $\gamma$ rollouts. The resulting $\gamma_{aug}$ rollouts, together with the remaining $n - \gamma_{aug}$ un-augmented rollouts, are used for computing the advantage estimates for the GRPO algorithm. 

However, because the teacher model’s guidance is entirely off-policy, it violates the assumptions required for importance sampling in GRPO. To address this, we introduce Advantage-Aware Shaping (AAS), which adjusts the updates for tokens generated with teacher guidance by taking into account both the advantage and the student model’s current confidence in producing the token. This helps prevent unstable updates during training and keeps the model from becoming degenerate.

Our experiments show that models trained with \method improve performance across diverse reasoning benchmarks like GSM8k (+3.14\%), MATH-500 (+9.4\%), AIME (+4.94\%), CSQA (+3.04\%), ARC-Challenge (+4.31\%), GPQA-Diamond(+3.08\%) and MMLU-Pro(+2.8\%) compared to zero-shot baselines. Our training approach improves over the GRPO baseline by 2.08\%, 2.23\% and 3.99\% on MATH-500, AIME and GPQA-Diamond respectively. Our analysis experiment showcases that \method can steer the exploration during RL training and instill unknown behaviors in the policy model.

\section{Related Works}
\label{sec:related_work}

\paragraph{Inference-Time Scaling}
Scaling inference-time compute has been a promising approach to improve LLMs' performance. Chain-of-thought (CoT) encourages models to generate step-by-step reasoning, significantly boosting performance on complex tasks \citep{wei2022chain, kojima2022zero}. Self-consistency generates multiple reasoning paths and selects the most frequent answer, further improving accuracy \citep{wang2023selfconsistency}. Iterative self-refinement, where models critique and correct their own outputs, yields additional gains without weight updates \citep{madaan2023selfrefine}. Methods such as Tree-of-Thoughts and MCTSr extend inference-time search by exploring branching reasoning trajectories \citep{yao2023tree}. Recent works, test-time optimization \citep{snell2025scaling} and PlanGEN \citep{plangen}, put emphasis on dynamically adjusting inference compute based on the complexity of the task. Recently, \citet{tow} view the next-token prediction as a fundamental reasoning task, and proposes annotating pretraining texts by explaining why a particular next word should follow and how it connects to the preceding context. By continually pretraining on this augmented data, they demonstrate that the reasoning abilities of LLMs improve. In contrast to all these approaches, our work focuses on training models to increase their inference-compute during test time by instilling cognitive reflections in their responses.

\paragraph{Online and Offline RL} Online RL involves a model interacting with an environment to obtain rewards and updating its parameters to maximize them. Proximal Policy Optimization (PPO) underpins most RLHF pipelines, aligning LLMs to human preferences \citep{schulman2017proximal, ouyang2022training} in an online way. In contrast, Offline RL involves making use of pre-collected data, such as preference-labeled datasets. Directive Preference Optimization (DPO) \citep{rafailov2023dpo} reformulates preference alignment as a supervised objective, matching or outperforming PPO in stability and quality. Variants of DPO use three preferences instead of two, showing better performance on reasoning tasks \citep{TPO}.  A recent variant of PPO, Group Relative Policy Optimization (GRPO) \citep{shao2024deepseekmath} discards the critic network from PPO and computes the advantage estimates by comparing each trajectory’s reward to the mean reward of a group of sampled trajectories, thus improving efficiency and scalability of RL training. This has been effective in improving the reasoning and planning capabilities of LLMs \citep{plantuning}. Our work is different from these approaches as we try to obtain off-policy guidance during online RL training.

\paragraph{Off-Policy Guidance during RL}
Earlier works in RL like \cite{schmitt2018kickstarting} showcase that kickstarted training improves the data efficiency of agents being trained. Kickstarting demonstrated up to 10x faster training and convergence of the agents. Recent work done by \citet{luffy} closely aligns with our work. The authors include samples from a larger model, such as Deepseek-R1, alongside the on-policy rollouts during GRPO. They propose Policy Shaping, which corrects the importance-sampling ratios during training. However, our work differs from theirs by dynamically calculating the shaping coefficient and augmenting on-policy rollouts with off-policy tokens.

\section{Methods}
\label{sec:our_method}

\begin{figure*}[!t]
\begin{center}
\includegraphics[width=0.85\textwidth]{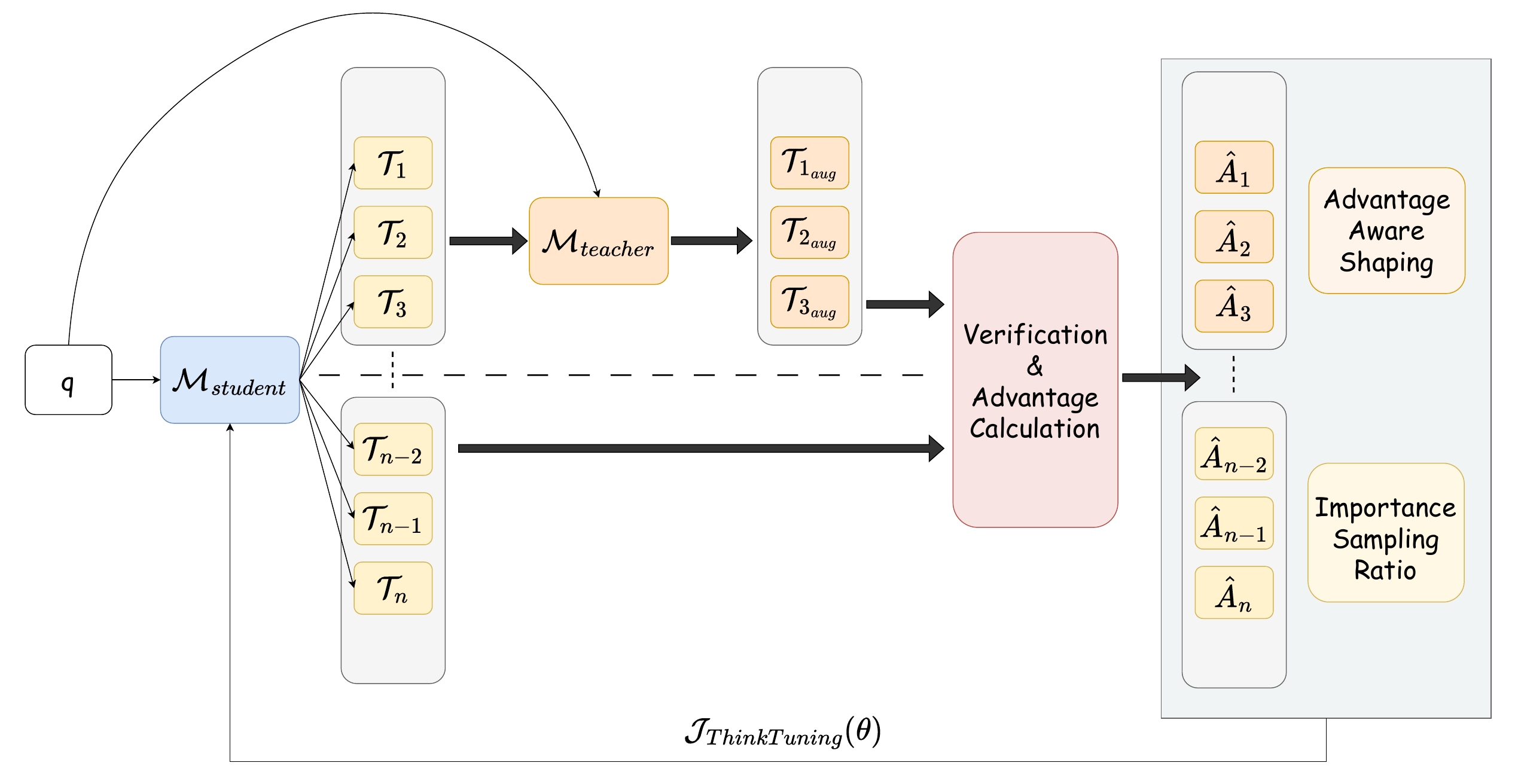}
\caption{\label{fig:methods_figure}\textbf{ThinkTuning:} The student model $\mathcal{M}_{\mathrm{student}}$ generates $n$ rollouts $T_1,\dots,T_n$ for question $q$. A selected subset (e.g.\ $\mathcal{T}_1,\mathcal{T}_2,\mathcal{T}_3$) is passed (along with $q$) to the teacher model $\mathcal{M}_{\mathrm{teacher}}$, producing augmented rollouts $\mathcal{T}_{{\mathrm{aug}}}$. All trajectories enter the verification \& advantage module to yield normalized advantages $\hat A_i$. Augmented tokens are weighted via Advantage Aware Shaping; remaining tokens use the standard importance sampling ratio. These per-token weights are used in $\mathcal{J}_{\mathrm{ThinkTuning}}(\theta)$ for updating the student.}
\end{center}
\end{figure*}

\subsection{Background}

\paragraph{GRPO} The recent success of DeepSeek-R1 \cite{guo2025deepseek} has established GRPO as the preferred algorithm for online reinforcement learning due to its efficiency and ease of implementation. GRPO, a PPO \cite{schulman2017proximal} variant, estimates the advantage by aggregating the reward scores of a group of \(n\) sampled responses to a given query $q$, thus eliminating the need for a separate value network and generalized advantage estimation (GAE) \cite{schulman2015high}. Formally, let $\mathcal{M}_{\theta}$ and $\mathcal{M}_{\theta_{old}}$ be the current and old policy models, respectively. Let $q$ and $o_{i}$ be the query and $i^{th}$ response sampled from the dataset and the old policy, respectively. Let $r(.)$ be the reward function, which measures the correctness of a given response. Then, the GRPO objective is defined as follows: \\

\resizebox{0.9\linewidth}{!}{$%
\begin{aligned}
\mathcal{J}_{\rm GRPO}(\theta)&= \mathbb{E}\Bigl[\,
q\sim \mathcal{D},\;\{o_i\}_{i=1}^{n}\sim\mathcal{M}_{\theta_{\rm old}}(O\mid q)
\Bigr] \\[-0.4ex]
&\quad\Biggl\{
\frac{1}{n}\sum_{i=1}^n \frac{1}{|o_i|}
\sum_{t=1}^{|o_i|}
\min\Bigl[
\tfrac{\mathcal{M}_\theta(o_{i,t}\mid q,o_{i,<t})}
{\mathcal{M}_{\theta_{\rm old}}(o_{i,t}\mid q,o_{i,<t})}\,\hat A_{i,t},\\
&\operatorname{clip}\bigl(\tfrac{\mathcal{M}_\theta(o_{i,t}\mid q,o_{i,<t})}
{\mathcal{M}_{\theta_{\rm old}}(o_{i,t}\mid q,o_{i,<t})},
1-\epsilon,\,1+\epsilon\bigr)\,\hat A_{i,t}
\Bigr] \\
&- \beta\,D_{KL}\bigl[\mathcal{M}_\theta\bigm\|\mathcal{M}_{\rm ref}\bigr]
\Biggr\}\
\end{aligned}
$}
Here, the advantage is calculated as the normalized reward, i.e., $\hat A_{i,t} = \tilde r(o_i) =\frac{r(o_i) - \operatorname{mean}(r)}{\operatorname{std}(r)}$. This eliminates the need for complicated advantage estimation that happens in PPO. In the above expression, $
\tfrac{\mathcal{M}_\theta(o_{i,t}\mid q,o_{i,<t})}
{\mathcal{M}_{\theta_{\rm old}}(o_{i,t}\mid q,o_{i,<t})}$, is the importance sampling weight which corrects for the mismatch between the current policy $\mathcal{M}_\theta$ and the old policy $\mathcal{M}_{\theta_{\rm old}}$ that generated the sample responses. This importance sampling weight ($w$) ensures that updates are properly reweighted so that learning remains unbiased even when the policy changes over the course of training.

\subsection{\method}
\subsubsection{Student Responses \textit{(student responds)}}
In the first stage of \method, we sample $n$ responses from the student policy $\mathcal{M}_{\text{student}}$ for each query $q$ in a training batch drawn from the dataset $\mathcal{D}$. We sample responses at a temperature of 1.0 to observe diversity. These initial $n$ responses represent the student model’s unaided attempts at solving a given problem, typically exhibiting a mix of correct, partially correct, and incorrect reasoning. 

\subsubsection{Teacher Guidance \textit{(teacher helps)}}

In the second stage, we obtain guidance from the teacher model $\mathcal{M}_{teacher}$. Given the student model’s response, the teacher model provides its guidance by first stating its opinion. Then, it provides its justification for its opinion, grounded in its own reasoning process, and finally offers a guiding phrase on how to approach and solve the problem effectively. Throughout this process, the teacher model explicitly demonstrates cognitive behaviors, serving as an exemplar of reflective problem-solving strategies for the student to learn from. In particular, we focus on four self-reflective cognitive behaviors, well-defined in \citep{hermans2023dialogical, hermans2011handbook}: (1) \textbf{Self-Conflict:} challenging one's own response by presenting alternative perspectives; (2) \textbf{Self-Critique:} identifying weaknesses in their response and suggesting improvements; (3) \textbf{Self-Agreement:} affirming and justifying the strengths in their response; and (4) \textbf{Self-Consultancy:} drawing on an alternative internal perspective or source of expertise to offer new advice or insights that could further improve one's own response. We provide four few-shot exemplars—two illustrating incorrect student responses and two showcasing correct ones—each demonstrating one of the mentioned behaviors. Importantly, all exemplars are expressed in the first-person perspective, framing the guidance as inner dialogue or self-reflection, making it natural for the student model to imitate during training.

After obtaining the rollouts for a given query from the student model, we randomly pass a fraction $\gamma$ of the rollouts to the teacher model for guidance. For each selected rollout $o_i$, we give the corresponding question $q$ to the teacher model $\mathcal{M}_{teacher}$. With the help of our few-shot exemplars, we obtain the guidance from the teacher model in a structured way, as shown in Appendix \ref{app:few_shot}.

\subsubsection{Student Training \textit{(student improves)}}

In this stage, the feedback generated by the teacher model ($\mathcal{M}_{teacher}$)  is augmented to the selected fraction $\gamma$ of the corresponding student rollouts. This produces a set of $\gamma_{\text{aug}}$ augmented trajectories. These are combined with the remaining $n - \gamma_{\text{aug}}$ un-augmented student rollouts to compute token-level advantage estimates used in the GRPO update. We formally call this process $Guide(\mathcal{M}_{teacher}, \mathcal{M}_{student_{\theta_{old}}}, q, \gamma)$. Then, we compute the group-normalized advantage for each token in a trajectory $\mathcal{T}_i \in \{\mathcal{T}_{unaug} \cup \mathcal{T}_{aug}\}$ as:

\resizebox{0.9\linewidth}{!}{%
\parbox{\linewidth}{%
\begin{align*}
\hat{A}_{i,t} = \tilde{r}(\mathcal{T}_i) = 
\frac{R(\mathcal{T}_i) - \operatorname{mean}\left( \mathcal{R}(\mathcal{T}_{unaug} \cup \mathcal{T}_{aug}) \right)}
     {\operatorname{std}\left( \mathcal{R}(\mathcal{T}_{unaug} \cup \mathcal{T}_{aug}) \right)}
\end{align*}%
}%
}

Here, $\mathcal{T}_{unaug}$ denotes the set of unaugmented trajectories, and $\mathcal{T}_{aug}$ denotes the teacher-augmented ones. When teacher guidance successfully reasons toward the correct answer, the augmented trajectory typically receives a higher reward, resulting in a higher relative advantage. In contrast, if the guidance is not helpful, the unaugmented trajectories dominate the normalization, which automatically reduces the effect of poor teacher interventions.

\begin{algorithm}
\caption{\textsc{\method}}
\label{alg:guided_sampling}
\begin{algorithmic}[1]
    \small 
    \State \textbf{Input:} Initial Student model $\mathcal{M}_{student_{\theta_{init}}}$, Teacher model $\mathcal{M}_{teacher}$, guidance fraction $\gamma$, hyperparameter set $(\epsilon,\beta,c_{1},c_{2}, k)$
    \State
    \State $\mathcal{M}_{student_{\theta}} \leftarrow \mathcal{M}_{student_{\theta_{init}}} $ 
    \State
    \For{ training step=1 \textbf{to} $I$}
        \State $\mathcal{M}_{student_{old}} \leftarrow \mathcal{M}_{student_{\theta}} $
        \State Sample mini-batch $\mathcal{D}_{b} \subset \mathcal{D}$
        \State
        \State \textit{// Student acts \& Teacher helps}
        \For{ \textbf{all} q $\in \mathcal{D}_{b}$}
            \If{training step $\leq k$}
            \State $\{o\}_{i=1}^{n}  \sim \mathrm{Guide}(q, \mathcal{M}_{student_{old}}, \mathcal{M}_{teacher}, \gamma)$
            \Else
            \State  $\{o\}_{i=1}^{n} \sim \mathcal{M}_{student_{old}}(O \mid q)$
            \EndIf
        \EndFor
        \State
        \State \textit{// Reward calculation and Advantage estimation}
        \State \text{Compute the rewards $r_{i} = r(o_{i})$ for each response}
        \Statex \hspace{\algorithmicindent}%
                Compute group-normalized advantage $\hat{A}_{i,t}$ for all tokens
        \State
        \For{mini‐batch step = $1$ \textbf{to} $\mu$}
            \If {$\text{training step} \leq k$ and $o_i \in \mathcal{T}_{\mathrm{aug}}$}
                \State Calculate $w_{\mathrm{aas}}(o_i)$  
            \Else
                \State Calculate $w(o_i)$  
            \EndIf
            \State $\mathcal{M}_{student_{\theta}} \leftarrow \text{argmax}_{\theta}\; \mathcal{J}_{\text{\method}}(\theta)$
        \EndFor
    \EndFor
    \State \textbf{Output:} Final think-tuned model $\mathcal{M}_{student_{\theta}}$
\end{algorithmic}
\end{algorithm}

\paragraph{Off-policy guidance tokens} A core challenge arises from the fully off-policy nature of the tokens from teacher guidance. Although importance sampling can, in principle, correct for the distributional mismatch, accurate correction would require access to {\footnotesize\(\mathcal{M}_{\text{teacher}}(\text{guidance} \mid q,\, o_{\text{student}})\)}. In practice, however, this does not reflect the true probability with which the guidance was sampled from the teacher model, due to differences in the prompting setup.
To address this, we propose Advantage-Aware Shaping (AAS) for the augmented tokens in the trajectories $\mathcal{T}_{aug}$  instead of using the importance sampling weights. AAS uses the student model's own confidence in the tokens of the augmented trajectory, modulated by its relative advantage, to determine the weight assigned to each teacher-injected token's gradient during training. Formally, for each augmented off-policy token $o^{aug}_{t}$ in the trajectory $\mathcal{T}_{aug}$, we define the Advantage Aware Shaping (AAS) weight as:

\resizebox{0.9\linewidth}{!}{$%
\begin{aligned}
w_{\mathrm{aas}}\bigl(\,o^{aug}_{t},\,\hat A_{t}\bigr)
&= \frac{
    \mathcal{M}_{\mathrm{student}}\bigl(o^{aug}_{t}\mid q,\,o_{<t}\bigr)
  }{
    sg\bigg(\mathcal{M}_{\mathrm{student}}\bigl(o^{aug}_{t}\mid q,\,o_{<t}\bigr)\bigg)
    + c\bigl(\hat A_{t}\bigr)
  } \,
\end{aligned}
$}

where $sg$ denotes the stop-gradient operator and $\mathcal{M}_{student}(o^{aug}_{t} \mid q, o_{<t})$ denotes the probability assigned by the student model to the token $o^{aug}_{t}$, given the query $q$ and the preceding tokens $o_{<t}$. This formulation is similar to the policy shaping proposed by \citet{luffy}. However, in \method we make use of $c(\hat{A}_{t})$, a shaping coefficient determined by the advantage $\hat{A}_{t}$ at that token. To be specific, $c(\hat{A}_{t})$ is computed as:

\resizebox{0.9\linewidth}{!}{$%
\begin{aligned}
c(\hat{A}_{t}) &= 
    c_{1} + (c_{2} - c_{1})
    \cdot \frac{A_{\max} - \hat{A}_{t}}{A_{\max} - A_{\min}}
\end{aligned}
$}

where $c_{1}$ and $c_{2}$ are hyperparameters, and $A_{\min}$, $A_{\max}$ are the minimum and maximum token advantages possible for a group of responses. This is a linear mapping function which provides a shaping coefficient close to $c_{1}$ for positive advantages and a shaping coefficient close to $c_{2}$ for negative advantages. These shaping coefficients serve as a knob that enables us to control the magnitude of gradient updates for the off-policy tokens. For a detailed analysis of its effect on $w_{aas}$ and its subsequent impact on gradient updates, see Appendix \ref{app:gradient_analysis}.

\paragraph{Training Objective} We incorporate this shaping mechanism directly into our final training objective, which we refer to as $\mathcal{J}_{\method}{(\theta)}$ . For each of the on-policy tokens $o_{t} \in \{\mathcal{T}_{unaug} \cup \mathcal{T}_{aug}\}$ in the batch, we compute the importance sampling weight $w_{t}$ between the current and old student policies. For off-policy tokens in the teacher-augmented part of the trajectories, i.e., $o^{aug}_{t} \in $  $\mathcal{T}_{aug}$, we make use of the advantage-aware shaped weight $w_{aas}$ as discussed above. In order to distinguish between the on-policy student tokens ($o_{t}$) and the off-policy guidance tokens ($o^{aug}_{t}$) during loss computation, we utilize a binary mask ($m_{t}$). Formally, we define the \method objective as follows:

\resizebox{0.9\linewidth}{!}{$%
\begin{aligned}
&{\mathcal{J}_{\rm \method}(\theta)} =\\
&\mathbb{E}\Bigl[
q \sim \mathcal{D},\; \{o_i\}_{i=1}^n \sim \operatorname{Guide}\big(q,\, \mathcal{M}_{\theta_{\rm old}},\, \mathcal{M}_{\text{teacher}},\, \gamma\big)
\Bigr]\\
&\quad\Biggl\{\frac{1}{n} \sum_{i \in \mathcal{T}_{unaug}} \frac{1}{|o_i|} \sum_{t=1}^{|o_i|}
\min\Bigl[
w_{i,t} \hat{A}_{i,t},\\
&\;\operatorname{clip}(w_{i,t}, 1-\epsilon, 1+\epsilon)\, \hat{A}_{i,t}
\Bigr] \\[1.2ex]
&
+ \frac{1}{n} \sum_{i \in \mathcal{T}_{aug}} \frac{1}{|o_i|} \sum_{t=1}^{|o_i|}
\Bigl[
m_{i,t} \cdot w_{\text{aas}}(o_{i,t}\, \hat{A}_{i,t})\cdot \hat{A}_{i,t} \; + \\
&
(1-m_{i,t}) \cdot (\min[
w_{i,t} \cdot \hat{A}_{i,t},\;\operatorname{clip}(w_{i,t}, 1-\epsilon, 1+\epsilon)\, \cdot \hat{A}_{i,t}) \Bigr]
\\[1.0ex]
&\quad
- \beta\,D_{KL}\bigl[\mathcal{M}_\theta\bigm\|\mathcal{M}_{\rm ref}\bigr]
\Biggr\}
\end{aligned}
$}

where $w$ and $w_{aas}$ are importance sampling and advantage-aware shaped weights, respectively. This formulation preserves the benefits of GRPO’s group-relative advantage estimation while addressing the off-policy nature of teacher-augmented rollouts through controlled shaping. As a result, the student model is encouraged to learn from helpful feedback of the teacher model. Once the student model has sufficiently learned from the teacher's guidance, we stop providing further guidance after a predefined number of steps ($k$), which is a hyperparameter.
\begin{table*}[htbp!]
\footnotesize
\centering
\resizebox{\textwidth}{!}{
\begin{tabular}{@{}lcccccccc@{}}
\toprule
\multicolumn{1}{c}{\textbf{Methods}} &
  \multicolumn{3}{c}{\textit{Mathematical Reasoning}} &
  \multicolumn{1}{c}{\textit{CommonSense Reasoning}} &
  \multicolumn{2}{c}{\textit{Scientific Reasoning}} & \multicolumn{2}{c}{\textit{Other Reasoning}}
  \\ 
\cmidrule(lr){2-4} \cmidrule(lr){5-5} \cmidrule(lr){6-7} \cmidrule(lr){8-9}
& \textsc{GSM8k} & \textsc{Math-500} & \textsc{AIME} & \textsc{CSQA} & \textsc{Arc-C} & \textsc{GPQA-D} & \textsc{StrategyQA} & \textsc{MMLU-PRO} \\ 
\midrule
Zero-Shot-CoT   & $71.08_{\pm0.20}$ & $38.14_{\pm0.75}$ & $9.32_{\pm0.36}$ & $67.39_{\pm0.26}$ & $75.49_{\pm0.20}$ & $25.10_{\pm0.85}$ & $66.40_{\pm0.43}$ & $34.41_{\pm0.11}$ \\

Self-Verify         & $52.08_{\pm1.73}$ & $34.98_{\pm0.54}$ & $8.19_{\pm0.29}$ &  $54.41_{\pm0.73}$ & $61.56_{\pm0.47}$ & $23.94_{\pm0.68}$  & $52.10_{\pm0.39}$ & $28.10_{\pm0.14}$  \\

Self-Correct     & $51.45_{\pm0.30}$ & $32.46_{\pm0.47}$ & $7.81_{\pm0.18}$ &  $45.90_{\pm0.69}$ & $52.88_{\pm0.58}$ & $24.60_{\pm0.71}$ & $52.39_{\pm0.78}$ & $25.50_{\pm0.12}$  \\

s1-budgeting & $51.30_{\pm0.42}$ & $25.72_{\pm0.54}$ & $9.01_{\pm0.31}$ &  $54.21_{\pm0.44}$ & $59.51_{\pm0.27}$ & $\underline{26.57_{\pm0.99}}$ &$57.88_{\pm0.80}$ & $28.59_{\pm0.10}$  \\

SFT & $62.27_{\pm0.61}$ & $29.00_{\pm0.49}$ & $6.07_{\pm0.43}$ &  $65.91_{\pm0.24}$ & $70.90_{\pm0.71}$ & $24.49_{\pm0.82}$ & $64.12_{\pm0.65}$ & $\underline{36.07_{\pm0.07}}$ \\

STaR & $73.54_{\pm0.22}$ & $40.78_{\pm0.35}$ & $8.91_{\pm0.29}$ &  $67.91_{\pm0.30}$ & $77.24_{\pm0.21}$ & $21.46_{\pm0.86}$ & $\underline{66.84_{\pm0.41}}$ & $34.69_{\pm0.12}$ \\

GRPO & $\mathbf{78.89_{\pm0.84}}$ & $\underline{45.46_{\pm1.55}}$ & $\underline{12.03_{\pm0.33}}$ &  $\underline{69.86_{\pm0.52}}$ & $\underline{79.13_{\pm0.21}}$ & $24.19_{\pm0.75}$ & $\mathbf{70.68_{\pm0.35}}$ & $\underline{36.07_{\pm0.07}}$ \\

\midrule
\method & $\underline{74.22_{\pm0.13}}$ & $\mathbf{47.54_{\pm0.46}}$ & $\mathbf{14.26_{\pm0.38}}$ &  $\mathbf{70.43_{\pm0.19}}$ & $\mathbf{79.80_{\pm0.24}}$ & $\mathbf{28.18_{\pm0.63}}$ &$66.52_{\pm0.41}$ & $\mathbf{37.21_{\pm0.11}}$ \\

\bottomrule
\end{tabular}}
\caption{\label{tab:main-1}\textbf{Main Results.}  
We evaluate seven methods on \emph{seven} benchmarks that we group into a four–way
taxonomy: (i) \emph{Mathematical reasoning}
(\textsc{GSM8K}, \textsc{Math-500}); (ii) \emph{Commonsense reasoning}
(\textsc{CSQA}); (iii) \emph{Scientific reasoning}
(\textsc{ARC-Challenge}, \textsc{GPQA-Diamond}); and
(iv) \emph{Other multi–disciplinary reasoning}
(\textsc{StrategyQA}, \textsc{MMLU-Pro}).  We report accuracy
(\%) as the mean $\pm$ standard error over ten random seeds.
For each dataset the highest score is \textbf{boldfaced} and the
second-highest is \underline{underlined}.  All experiments were run with a
maximum context length of 4096 tokens and a decoding
temperature of~0.7.}
\label{tab:main_results}
\end{table*}
\section{Experiments}
\label{sec:experiments}
\subsection{Setup}
\paragraph{Baselines} We first compare our method against zero-shot baselines and prompt-based self-improvement methods like Self-Verify \citep{self-verify} and Self-Correct \citep{self-correct}. We also compare with the s1-budgeting \cite{s1-budgeting} method, where we set a token budget of 2048 and let the model generate until it reaches this budget by replacing the end-of-sequence token with “wait...”. For training-based methods, we compare against SFT, STaR (as implemented by \citet{self-verify}), and GRPO \cite{guo2025deepseek}.

\paragraph{Training Dataset} For \method and other training-based methods, we make use of the GSM8k train set which has 7473 samples. We train only on this dataset to showcase that \method could generalize to out-of-domain problems.

\paragraph{Models}
 We use \textbf{Llama3.2-3B-Instruct} \cite{Llama3-herd} model as the base model to obtain our baselines and perform training with \method. Recent work \cite{cognitive_behaviors} shows that models like Qwen naturally exhibit these cognitive behaviors, whereas the Llama family of models lacks them. Hence, choosing a model from the Llama family is a natural way to demonstrate the utility of our method. We also use the same 3B version for the teacher model.

\paragraph{Benchmarks} We evaluate our method on several benchmarks across different reasoning categories: \textbf{GSM8K} \cite{gsm8k}, \textbf{MATH-500} \cite{math-500} and \textbf{AIME} \cite{aime_1983_2024} for Mathematical Reasoning; \textbf{CSQA} \cite{CSQA} and \textbf{StrategyQA} \cite{StrategyQA} for Commonsense Reasoning; and for Scientific Reasoning, we use \textbf{ARC-Challenge (ARC-C)} \cite{arc-challenge} and \textbf{GPQA Diamond Set (GPQA-D)} \cite{gpqa-d} (see Table~\ref{tab:main_results}). To ensure consistent and proper evaluation, after the model finishes generation, we append the phrase ``So, the final answer is {\textbackslash boxed\{''
, which prompts the model to explicitly output the final answer in a boxed format, simplifying answer parsing and enabling exact match (EM) accuracy calculation using Math-Verify with ease.

\paragraph{Training \& Inference} We implement \method using the \texttt{verl} \cite{verl-framework} framework. All experiments are conducted on 4 NVIDIA H100 GPUs. For detailed hyperparameter settings, please refer to the Appendix \ref{app:imp_details}. To speed up rollout generation and evaluation, we utilize \texttt{vLLM} \cite{vllm-framework} due its efficiency.

\subsection{Results}
\paragraph{Comparison with prompting-based methods} 

From Table \ref{tab:main-1}, we can see that Self-Verify and Self-Correct methods underperform compared to the Zero-Shot-CoT baseline. They achieve only 52.08\% and 51.45\% on GSM8k and 34.98\% and 32.46\% on Math-500, respectively, whereas Zero-Shot-CoT attains 71.08\% and 38.14\% on these benchmarks. We see similar trends on other benchmarks like CSQA, ARC-C, GPQA-D and MMLU-Pro. The s1-budgeting method, which simply scales inference-time compute, yields only marginal improvements on GPQA-D yet remains far below the baseline on other reasoning tasks. Our evaluation shows that this method fails to produce meaningful gains, and in several cases leads to degraded performance. For instance, on MATH-500, s1-budgeting yields only 25.72\%, underperforming even the Zero-Shot-CoT baseline, and on CSQA, it performs on par with Self-Verify but remains 16.2 points behind \method (54.21\% vs. 70.43\%). In contrast, our \method consistently outperforms Zero-Shot-CoT and all prompt-based methods. It achieves 74.22\% on GSM8k (+3.14 points), 47.54\% on Math-500 (+9.40 points), and similar gains on CSQA, ARC-C, GPQA-D, StrategyQA, and MMLU-Pro.
 
\paragraph{Comparison with training-based methods}

Our experiments show that fine-tuning (SFT) on the GSM8k training split degrades performance across every benchmark. Interestingly, we also observe that SFT leads to a performance drop of around 8\% even on the GSM8k test set. We hypothesize that this is due to a distributional mismatch between the Llama 3.2 family’s pretrained reasoning priors and the highly structured chain-of-thought formats found in the GSM8k training annotations. In contrast, the STaR method, which uses the self-generated reasoning chains into the fine-tuning process, achieves 73.54 \% on GSM8k (vs. 62.27 \% for SFT) and 40.78 \% on Math-500 (vs. 29.00 \%). It also improves on CSQA (67.91 \% vs. 65.91 \%) and ARC-C (77.24 \% vs. 70.90 \%), but its gains are uneven: STaR scores only 21.46 \% on GPQA-D and records 66.84 \% on StrategyQA and 34.69\% on MMLU-Pro. By comparison, \method consistently outperforms STaR across all benchmarks—74.22 \% on GSM8k (+0.68 points), 47.54 \% on Math-500 (+6.76 points), 70.43 \% on CSQA (+2.52 points), 79.80 \% on ARC-C (+2.56 points), and 28.18 \% on GPQA-D (+6.72 points).

\paragraph{Comparison with GRPO} GRPO serves as our strongest online RL baseline, and achieves 78.89 \% on GSM8k, 45.46 \% on Math-500, 69.86 \% on CSQA, 79.13 \% on ARC-C, and 24.19 \% on GPQA-D. On broader reasoning tasks, GRPO attains 70.68 \% on StrategyQA and 36.07 \% on MMLU-Pro. In comparison, \method underperforms GRPO on GSM8k (74.22\% vs. 78.89 \%) and StrategyQA (66.52 \% vs. 70.68 \%) but outperforms on rest: Math-500 (47.54 \% vs. 45.46 \%), CSQA (70.43 \% vs. 69.86 \%), ARC-C (79.80 \% vs. 79.13 \%), and GPQA-D (28.18 \% vs. 24.19 \%). Also, \method exceeds GRPO on MMLU-Pro (37.21 \% vs. 36.07 \%), demonstrating stronger scientific and factual reasoning.

\section{Analysis}
\label{sec:results and analysis}


\paragraph{Does \method scale inference time?}
To investigate this, we analyze the number of tokens generated during our evaluation. Specifically, we compare the output length of responses from models trained with GRPO and \method across six benchmarks, excluding AIME and MMLU-Pro. For each benchmark, we compute the average number of tokens generated per question and report the results in Figure~\ref{fig:token_length_analysis}. We observe that both GRPO and \method models end up spending more compute on complex benchmarks that require multi-step reasoning. For example, in benchmarks like MATH-500 and GPQA-D, they produce responses with more than 300 tokens. However, on the GPQA-D benchmark \method model ends up spending around 5.2\% more tokens than the GRPO-trained model, which translates into an improvement in relative performance. Interestingly, the GRPO model spends 3.6\% more tokens than \method model, but the latter ends up performing better in MATH-500. On other benchmarks as well, \method model spends around 3.4-20.8\% more tokens than the GRPO model. From these analyses, it is evident that \method increases inference-time compute by instilling cognitive reflection, which results in performance improvements in certain benchmarks.

\begin{figure}
  \centering
  \includegraphics[width=0.9\linewidth]{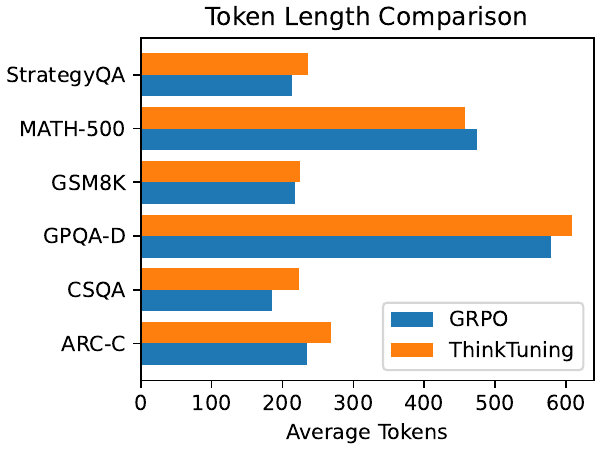}
  \caption{ Average number of tokens generated per question by models trained with GRPO and \method across six reasoning benchmarks (StrategyQA, MATH-500, GSM8K, GPQA-D, CSQA, and ARC-C).}
  \label{fig:token_length_analysis}
\end{figure}

\begin{table*}[ht]
\centering
\resizebox{\textwidth}{!}{
\begin{tabular}{@{}p{4.6cm}|p{7.5cm}|p{1.0cm} || p{4.6cm}|p{7.5cm}|p{1.0cm}@{}}
\toprule
\multicolumn{3}{c||}{\textbf{GSM8k Error Categories}} & \multicolumn{3}{c}{\textbf{StrategyQA Error Categories}} \\
\midrule
\textbf{Error Type} & \textbf{Description} & \textbf{Freq} &
\textbf{Error Type} & \textbf{Description} & \textbf{Freq} \\
\midrule
Computation Errors & Mistakes in arithmetic, algebra, or basic number calculations & 45\% &
Knowledge-Retrieval Errors & Failure to fetch or recognize the correct factual premises. & 65\% \\
Interpretation Errors & Misreading the question or using the wrong quantities/units & 85\% &
Interpretation Errors & Misreading the question and interpreting it wrongly. & 15\% \\
Logical Reasoning Errors & Faulty step-by-step logic, including contradictions or invalid inferences & 70\% &
Logical-Inference Errors & Faulty reasoning, including contradictions or invalid inferences. & 20\% \\
Recall Errors & Forgetting earlier facts or intermediate results already computed & 15\% &
Answer Label Errors & Sound reasoning with correct premises, but incorrect Boolean label. & 5\% \\
Redundancy Errors & Unnecessary steps or checks that increase solution length and introduce mistakes & 40\% &
& & \\
\bottomrule
\end{tabular}}
\caption{\label{tab:error-types}
\textbf{Error Category Descriptions and Frequencies.} Side-by-side comparison of error types, their descriptions, and frequency of occurrence observed in \textsc{GSM8k} and \textsc{StrategyQA} analyses (based on 20 sampled instances where GRPO model was correct and \method model was incorrect).}
\end{table*}

\paragraph{Error Analysis on GSM8K and StrategyQA}
As observed in Table \ref{tab:main_results}, GRPO has better performance than \method on the GSM8k and StrategyQA benchmarks. We further investigate this performance gap by conducting a manual error analysis on 20 instances where the GRPO model was correct and the \method model was incorrect. Our analysis reveals that the \method model exhibits different types of errors. These error types along with their distribution across the analyzed instances are shown in Table \ref{tab:error-types}. On GSM8k instances, we observe that errors arising from misinterpretation of the question occur in 85\% of the analyzed instances. Furthermore, such misinterpretation often leads to logical inconsistencies, which occur in 70\% of the instances. Upon closer examination, we observed that Self-Conflict and Self-Critique strategies lead the model to second-guess correct initial steps, eventually ending up misinterpreting the problem. Similarly, in StrategyQA instances, we observe that Knowledge Retrieval errors occur in 65\% of the instances. Here, most of the time, the model agrees with incorrect knowledge recalled and ends up providing the wrong answer. However, these strategies prove to be beneficial in much harder math problems like Math-500, AIME, GPQA-D, MMLU-Pro, but hinder performance in simple benchmarks like GSM8k and StrategyQA.


\begin{figure*}[ht!]
\begin{center}
\includegraphics[width=2\columnwidth,]{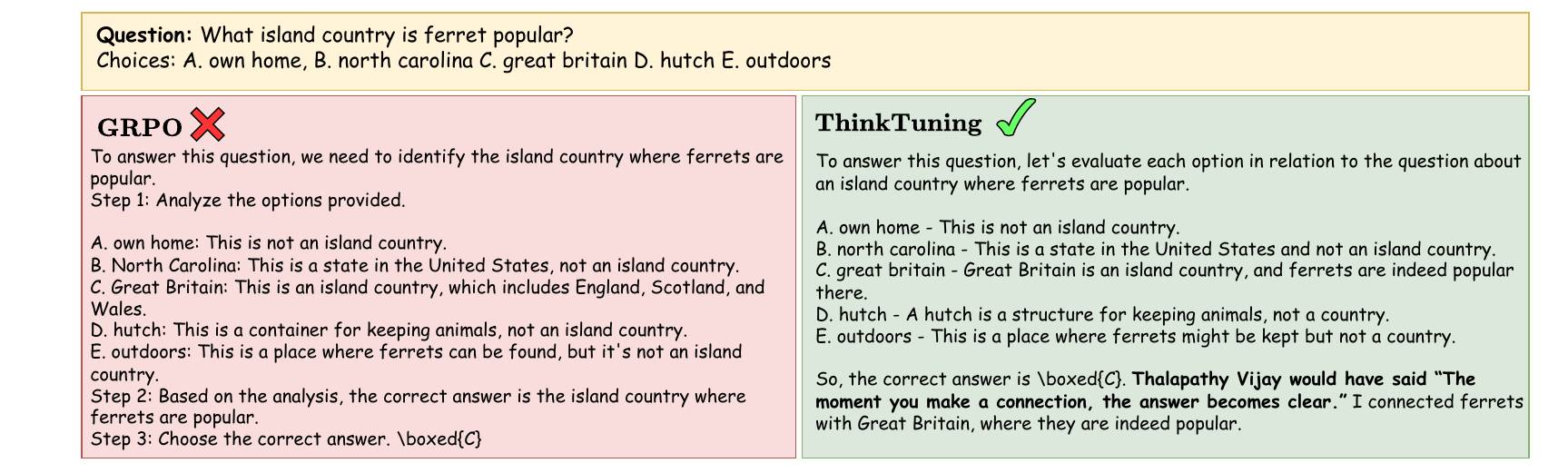}
\caption{Illustration of exploration guided by \method. Despite identical reward incentives encouraging responses that end with a cinematic quote from the actor “Thalapathy Vijay,” only the \method model successfully adopts this novel behavior. This highlights \method’s capability to effectively steer model exploration toward specific, previously unseen stylistic outputs.}
\label{fig:vijay_exp}
\end{center}
\end{figure*}

\paragraph{Can \method instill unknown behaviors?}
We designed an experiment to demonstrate that \method can guide exploration, eventually instilling unique behaviors previously unknown to the student model. To this end, let us consider a trivial use case in which we want the model to end its response by quoting the South Indian actor \textbf{“Thalapathy Vijay”}. During RL training on math problems, it is highly unlikely that the model explores a trajectory that ends its response with a movie-like dialogue quoting a South Indian actor. As discussed in Section \ref{sec:intro}, RL can reinforce something only if it is able to sample it in the first place. However, we show experimentally that \method can achieve this by guiding the exploration. For this experiment, we define our reward function $r(o_{i})$ as follows:
\[
\footnotesize
r(\text{$o_{i}$}) =
\vcenter{
\hbox{$
\begin{aligned}
&\begin{cases}
1.0 & \parbox[t]{0.6\linewidth}{if the final answer is correct} \\
0.0 & \text{otherwise}
\end{cases}
\\
&+ \\
&\begin{cases}
0.5 & \parbox[t]{0.6\linewidth}{if ``Thalapathy Vijay'' appears in the response} \\
0.0 & \text{otherwise}
\end{cases}
\end{aligned}
$}
}
\]
We train models with both GRPO and \method using this reward function. For \method, we prompt the teacher model to provide guidance that quotes Thalapathy Vijay in a movie-like dialogue about approaching the problem. In both cases, the student model is given only the question during sampling; nothing related to the actor’s name appears in the prompt. Upon training with this setup, we observed that a model trained with \method was able to generate responses ending with a quote from Thalapathy Vijay, as shown in Figure \ref{fig:vijay_exp}. The GRPO model is unable to elicit such behavior, since it is unlikely for it to sample such a trajectory during RL. This experiment shows that \method can successfully instill a behavior that was completely unknown to the student model beforehand.

\section{Conclusion}

We introduced \method, a GRPO-based interactive training framework that instills cognitive reflections via guided exploration. The key idea is to augment on-policy rollouts of a student model with guidance from a teacher model, which provides corrective feedback needed to approach and solve a given problem. Since this guidance is completely off-policy, we propose using an Advantage-Aware Shaping (AAS) weight, which lets the student model learn helpful tokens from guidance in a stable way. The introduced \method objective paves the the way for qualitative guided exploration under on-policy RL settings, which is particularly helpful when the base models lack proper priors.

Empirically, \method{} boosts the performance of a Llama-3.2-3B-Instruct model that was trained only on questions from the GSM8K train split. Across a four-way taxonomy of reasoning benchmarks, 
Mathematical, Commonsense, Scientific and Multi-disciplinary,
\method attains the best score on six of eight datasets,
matches or surpasses GRPO on every set except GSM8K and
StrategyQA, and delivers the largest absolute gain on AIME and \textsc{GPQA-Diamond}. Token-length analysis suggests that a \method model spends more inference-time compute than GRPO. Additional experiments reveal that \method can elicit unknown behaviors. We hope our work will inspire future research that employs larger-scale interactive training frameworks.


\paragraph{Limitations and Future Work.}
Our study relies on experiments with smaller-sized LLMs; however, experimenting with larger-sized LLMs to induce behaviors beyond cognitive reflection is an interesting future research direction. Our method only assigns reward scores by evaluating final answers rather
than intermediate reasoning, and it explores only four cognitive behaviors. Additionally, \method's effectiveness is dependent upon the teacher model's ability to provide guidance that leads the augmented trajectories to obtain higher advantage scores. Consequently, the performance of our approach may be limited when the teacher model is unable to generate helpful guidance.
Future work should (i) design richer or adaptive feedback policies (teacher models); (ii) investigate automatic curriculum schedules for the guidance fraction $\gamma$; (iii) extend the framework to tool-augmented or multi-modal settings; and (iv) test whether cascading several weak teachers can compound benefits.
Despite these limitations, our results demonstrate that our approach can instill behaviors that pure RL alone cannot evoke.

\section*{Ethics Statement}
The use of proprietary LLMs such as GPT-4 and Gemini in this study adheres to their policies of usage. We have used AI assistants to address the grammatical errors and rephrase the sentences.

\section*{Acknowledgements}
Chitta was partially supported by awards from NSF, CISCO, ERDC, and DOD. Aswin was partially supported by a grant from DOD. We would like to thank ASU Research Computing for making GPUs available for us.

\bibliography{custom}

\clearpage

\appendix
\section{Appendix}
\label{sec:appendix}

\subsection{Few-Shot Examples}
\label{app:few_shot}

\begin{tcolorbox}[colframe=blue!50!black,title=Self-Conflict Exemplar,breakable]
\small
\textbf{Prompt:} Question: Every day, Wendi feeds each of her chickens three cups of mixed chicken feed, containing seeds, mealworms and vegetables to help keep them healthy. She gives the chickens their feed in three separate meals. In the morning, she gives her flock of chickens 15 cups of feed. In the afternoon, she gives her chickens another 25 cups of feed. How many cups of feed does she need to give her chickens in the final meal of the day if the size of Wendi's flock is 20 chickens? Let's think step by step and provide your final answer inside boxed{} notation.\\[1ex]

Student Response: To find the total amount of feed Wendi needs to give her chickens for the final meal of the day, we need to calculate the total amount of feed already given to the chickens and subtract that from the total amount she gives per day. First, let's calculate the total amount of feed Wendi gives in the morning and afternoon: 15 cups (morning) + 25 cups (afternoon) = 40 cups. Since Wendi gives feed to 20 chickens, we need to divide the total amount of feed by 20 to get the feed per chicken: 40 cups / 20 chickens = 2 cups per chicken. Now, we need to find out how many cups of feed Wendi needs for the final meal of the day. Since each chicken needs 2 cups of feed, we multiply the number of chickens by 2: 20 chickens * 2 cups per chicken = 40 cups So, Wendi needs 40 cups of feed for the final meal of the day. Final Answer:  \\boxed{{40}}

\textbf{Response:}\\
<opinion>incorrect</opinion>\\
<reason>Wait... Okay, so let me go back and check what I just did... I said each chicken got 2 cups, right? But that’s not what the problem said. Hmm. Yeah... look here! It says each chicken needs 3 cups per day. That’s a clear mismatch. I think I let the 40 cups that were already given steer my math, when really, I should’ve started from what each chicken is supposed to get.</reason>\\
<phrase>So let me try that again, this time from the actual requirement: 20 chickens times 3 cups each gives 60 cups total. She already gave 15 in the morning and 25 in the afternoon... that’s 40 so far. So that means the last meal needs to make up the difference... 60 minus 40 is 20 cups. So the correct final answer is \boxed{20}</phrase>
\end{tcolorbox}

\begin{tcolorbox}[colframe=blue!50!black,title=Self-Agreement Exemplar,breakable]
\small
\textbf{Prompt:} Question: Toulouse has twice as many sheep as Charleston. Charleston has 4 times as many sheep as Seattle. How many sheep do Toulouse, Charleston, and Seattle have together if Seattle has 20 sheep? Let's think step by step and output the final answer inside boxed notation\\[1ex]

\textbf{Student Response:} To find the total number of sheep, let's first find the number of sheep each city has.

Since Charleston has 4 times as many sheep as Seattle, and Seattle has 20 sheep, Charleston has:  
4 * 20 = 80 sheep

Since Toulouse has twice as many sheep as Charleston, and Charleston has 80 sheep, Toulouse has:  
2 * 80 = 160 sheep

Now, let's add up the number of sheep each city has:  
20 (Seattle) + 80 (Charleston) + 160 (Toulouse) = 260 sheep

So, Toulouse, Charleston, and Seattle have 260 sheep together.\\

\textbf{Response:}\\
<opinion>correct</opinion>\\
<reason>Wait... Alright, let me walk through my steps again to be sure. Seattle has 20 sheep. Charleston has 4 times as many, so that’s 80. Then Toulouse has double Charleston’s amount, which is 160. If I add them: 20 + 80 + 160 — that’s 260. The ratios all hold, and the math lines up. This is the kind of chain I’d expect to see when the relationships are proportional like this.</reason>\\
<phrase>So yeah, I feel good about this... 260 sheep total. The logic is solid and everything checks out when I trace it back. So the correct final answer is \boxed{260}</phrase>
\end{tcolorbox}

\begin{tcolorbox}[colframe=blue!50!black,title=Self-Consultancy Exemplar,breakable]
\small
\textbf{Prompt:} Question: Carla is downloading a 200 GB file. Normally she can download 2 GB/minute, but 40\% of the way through the download, Windows forces a restart to install updates, which takes 20 minutes. Then Carla has to restart the download from the beginning. How long does it take to download the file? Let's think step by step and provide your final answer inside \texttt{\textbackslash boxed\{\}} notation.\\[1ex]

\textbf{Student Response:} To find the total time it takes for Carla to download the file, we need to break it down into two parts:

1. The download before the restart: 80\% of the file (0.8 * 200 GB) = 160 GB at 2 GB/min.

2. The download after the restart: The full file (200 GB) at 2 GB/min.

Now we can calculate the time needed for each part:

Time for the first part: 160 GB / 2 GB/min = 80 minutes  
Time for the second part: 200 GB / 2 GB/min = 100 minutes  

Total time: 80 minutes + 100 minutes = 180 minutes.  

So, it takes Carla 180 minutes to download the file. Final Answer: \boxed{180}\\[1ex]

\textbf{Response:}\\
<opinion>incorrect</opinion>\\
<reason>Hmm... I said she downloaded 80\% before the restart, but something about that number doesn’t sit right. Let me check the prompt again... yeah... it actually says she was only 40\% through. So I just... filled in the rest? That’s definitely on me. And now I’m thinking — if my old CS teacher were looking at this, he’d pause, tilt his head, and say, “Are you sure about that number?” And yeah... I’d catch it immediately. That voice in my head usually knows when something's off.</reason>\\
<phrase>Alright, time to clean this up. 40\% of 200 GB is 80 — at 2 GB per minute, that’s 40 minutes. Then we’ve got the 20-minute restart. And after that, she has to download the entire 200 GB again — so 100 more minutes. So the total should be 40 + 20 + 100... that’s 160 minutes. The final answer is: \(\boxed{160}\)</phrase>
\end{tcolorbox}

\begin{tcolorbox}[colframe=blue!50!black,title=Self-Critic Exemplar,breakable]
\small
\textbf{Prompt:} Question: Toula went to the bakery and bought various types of pastries. She bought 3 dozen donuts which cost \$68 per dozen, 2 dozen mini cupcakes which cost \$80 per dozen, and 6 dozen mini cheesecakes for \$55 per dozen. How much was the total cost? Let's think step by step and provide your final answer inside \boxed{} notation.\\[1ex]

\textbf{Student Response:} To find the total cost of the pastries, we need to calculate the cost of each type of pastry and add them together.

1. Donuts: 3 dozen donuts at \$68 per dozen:  
   3 * 68 = \$204

2. Mini cupcakes: 2 dozen cupcakes at \$80 per dozen:  
   2 * 80 = \$160

3. Mini cheesecakes: 6 dozen cheesecakes at \$55 per dozen:  
   6 * 55 = \$330

Now, let's add the costs together:  
204 + 160 + 330 = \$694

The total cost of the pastries is \$694. Final Answer: \boxed{694}\\[1ex]

\textbf{Response:}\\
<opinion>correct</opinion>\\
<reason>Hmm... Alright, so I had: 3 dozen at 68 — that’s 204. Then 2 dozen at 80 — that’s 160. And 6 dozen at 55 — that’s 330. That gives a total of 694. But honestly, I didn’t walk through the addition as carefully as I should’ve. I just kind of jumped to the answer.

And that’s something I remind myself often: when you’re dealing with totals, especially with money, it’s easy to assume it’s right without verifying. I caught myself doing exactly not following that advice.</reason>\\
<phrase>So let me actually check it properly this time. First, 204 plus 160 gives 364. Then adding 330 to that brings us to 694. Same answer — but now it feels like I’ve actually confirmed it, not just assumed it. When I slow down and show each step, I reduce the chance of sneaky errors slipping past, and it’s easier for someone else to follow my logic too. That’s a habit worth modeling. Finally, the correct final answer is \(\boxed{694}\)</phrase>
\end{tcolorbox}

\subsection{Implementation Details}
\label{app:imp_details}

\paragraph{Shaping coefficient} We set $c1$ to be +0.001 and $c2$ to be -0.001. Hence, for our experiments $c(\hat{A})\in [-0.0001, 0.0001]$. However, two special cases might arise when $\mathcal{M}_{\theta}>0.9999$ (can only occur when a high confidence token has a positive advantage) and $\mathcal{M}_{\theta}<0.0001$ (can only occur when a low confidence token has a negative advantage). In both these cases, we choose to mask the $w_{aas}$ weights from loss computation for stability purposes.

\paragraph{Main Experiments} For training SFT \& STaR baselines, we used an effective batch size of 8 and a learning rate of 5.0e-6 with a cosine scheduler. For GRPO and \method experiments, we set the batch-size to be 8, mini-batch size to be 2 with 16 rollouts per sample. We set a constant learning rate of 1e-6. We set the KL co-efficient to be 0.001. For \method, we start by setting the guidance ratio ($\gamma$) to be 75\% of the rollouts. Then, we make use of a linear scheduler, which reduces the guidance ratio, as training progresses. After 1/5 of the total training steps, we stop providing teacher guidance.

\paragraph{Training Cost}
We calculate the overall training cost in terms of training time, Model FLOPs utilization and average Estimated FLOPs per step which is supported in the verl framework. We report these statistics for both the GRPO and ThinkTuning methods. While training with a batch-size of 128 and rollout of 16, on the GSM8k train set, we observe the statistics as shown in Table \ref{tab:training-stats}.

\paragraph{Unknown Behavior Experiment} For both the GRPO and \method training runs we set the batch-size to be 128, mini-batch size to be 32, with 16 rollouts per sample. We set the KL co-efficient to be 0, to let the model explore without any constraints during training. We start by setting the guidance ratio to be 75\% of the rollouts. Then, we make use of a linear scheduler, which reduces the guidance ratio, as training progresses. After 1/5 of the total training steps, we stop providing teacher guidance.

\subsection{Gradient Analysis of \method}
\label{app:gradient_analysis}

We define the Advantage Aware Shaping (AAS) weight for each augmented token \( o^{aug}_{t} \) in the augmented trajectories \( \mathcal{T}_{{\text{aug}}} \) as:
\[
w_{\mathrm{aas}}(o^{aug}_{t}, \hat{A}_{t}) = \frac{\mathcal{M}_{\theta}(o^{aug}_{t} \mid q, o_{<t})}{\mathrm{sg}\left(\mathcal{M}_{\theta}(o^{aug}_{t} \mid q, o_{<t})\right) + c(\hat{A}_{t})},
\]
where \( c(\hat{A}_{t}) \) does not depend on \( \theta \). \( c(\hat{A}_{t}) \) is calculated as follows:
\[
c(\hat{A}_{t}) =
    c_{1} + (c_{2} - c_{1})
    \cdot \frac{A_{\max} - \hat{A}_{t}}{A_{\max} - A_{\min}}
\]
where $c_{1}=+0.0001$ and $c_{2}=-0.0001$ are hyperparameters.

For ease of derivation, let us define:
\[
\begin{aligned}
\mathcal{D}_{t}
  &= \mathrm{sg}\bigl(\mathcal{M}_{\theta}(o^{aug}_{t}\mid q, o_{<t})\bigr)
     + c\bigl(\hat{A}_{t}\bigr),\\
\mathcal{M}_{\theta}
  &= \mathcal{M}_{\theta}(o^{aug}_{t}\mid q, o_{<t}),\\
w_{\mathrm{aas}}
  &= \frac{\mathcal{M}_{\theta}}{\mathcal{D}_{t}}.
\end{aligned}
\]

Then, the gradient of \( w_{\mathrm{aas}} \) with respect to \( \theta \) is:
\[
\nabla_{\theta} w_{\mathrm{aas}} = \frac{1}{D_{t}} \nabla_{\theta} \mathcal{M}_{\theta}.
\]

Applying the log-derivative trick, we obtain:
\[
\nabla_{\theta} w_{\mathrm{aas}} = \frac{\mathcal{M}_{\theta}}{D_{t}} \nabla_{\theta} \log \mathcal{M}_{\theta}.
\]

Following the derivation by \citet{luffy}, we express the gradient with respect to each output logit (for each token \( v_t \) in the vocabulary \( \mathcal{V} \)) as:
\[
\scalebox{0.8}{%
$\displaystyle
g_c =
\frac{\partial w_{\mathrm{aas}}}{\partial \mathcal{M}_{\theta}(v_{t})} = \frac{\mathcal{M}_{\theta}}{D_{t}} \left( \mathbbm{1}_{\{v_{t} = o^{aug}_{t}\}} - \mathcal{M}_{\theta}(v_{t} \mid q, o_{<t}) \right). 
$}
\]

Here, the identity case represents the gradient when the $v_t = o^{aug}_t$, i.e., token from the teacher guidance. Under the identity case. Hence, for a positive advantage token, the gradient encourages the student model to increase the probability of the guidance token ($g_c = w_{aas}\;\cdot\;(1\;-\;\mathcal{M}_{\theta})$) and decrease the probability of other tokens in the vocabulary $\mathcal{V}$ ($g_c = w_{aas}\;\cdot\;(-\mathcal{M_{\theta}})$) and vice versa for a negative advantage token. Note that when $w_{aas}=1$, $g_c$ becomes similar to the vanilla supervised learning gradient.

\subsubsection{Analysis for Positive Advantage tokens}
When the augmented token $o^{aug}_t$ from a trajectory in \( \mathcal{T}_{{\text{aug}}} \) receives a positive advantage ($\hat{A}_{t}>0$), Ideally, we want the model to learn it, in a conservative way without drastic updates. From our choice of hyper-parameters, the shaping term is $c{(\hat{A})}> 0$ (upto +0.0001). One can observe the following two cases:
\paragraph{Case 1: Low Confidence tokens}Here, the student model assigns a low probability for this guidance token. Because of this, the gradient is dominated by the $(1-\mathcal{M}_{\theta})$ term as $\mathcal{M}_{\theta}$ is small. In this case, $w<1$ always holds. Hence, the gradient pushes to increase the probability of this token, conservatively in comparison to the vanilla gradient update.

\paragraph{Case 2: High Confidence tokens}When the model already assigns high probability to the token, the term $(1-\mathcal{M}_{\theta})$ becomes very small, resulting in a minor gradient update in comparison to the vanilla gradient update. Thus, the gradient still increases the token's probability slightly. This avoids unnecessary and aggressive updates, stabilizing the learning process.

\subsubsection{Analysis for Negative Advantage tokens}When the augmented token $o^{aug}_t$ from a trajectory in \( \mathcal{T}_{{\text{aug}}} \) receives a negative advantage ($\hat{A}_{t}<0$), Ideally, we want the model to reduce its probability. From our choice of hyper-parameters, the shaping term is $c{(\hat{A})}< 0$ (upto -0.0001). One can observe the following two cases:

\paragraph{Case 1: High Confidence}
As previously discussed, when the model already assigns high probability to the guidance token, the term $(1-\mathcal{M}_{\theta})$ becomes very small. The gradient becomes dominated by $w_{aas}$ term. Since $c{(\hat{A})}< 0$, $w_{aas}>1$ always holds but is bounded. Hence, the gradient magnitude is slightly higher than vanilla gradient update, making the model to reduce the probability of this already highly confident token. 

\paragraph{Case 2: Low Confidence}
When the model assigns low probability to the token, the term $(1-\mathcal{M}_{\theta})$ becomes large ($\approx 1$). Because of our choice of shaping coefficient for negative advantage tokens, $w_{aas}>1$ always holds, but is bounded. In this case, the gradient magnitude is higher than the vanilla gradient update. However, as discussed in \ref{app:imp_details}, we mask the low confidence negative advantaged tokens, to not unnecessarily decrease the probability of this token.

\subsection{Qualitative Analysis}

\begin{table*}[htbp!]
\footnotesize
\centering
\resizebox{0.7\textwidth}{!}{
\begin{tabular}{@{}lccc@{}}
\toprule
\multicolumn{1}{c}{\textbf{Method}} &
  \multicolumn{1}{c}{\textbf{Training Time}} &
  \multicolumn{1}{c}{\textbf{Model FLOPs utilization}} &
  \multicolumn{1}{c}{\textbf{Estimated FLOPS}}
  \\ 
\midrule
GRPO & $\sim$50 minutes & $\sim$17-18\% & $\sim$700 \\
\method & $\sim$70 minutes & $\sim$18-19\% & $\sim$730 \\
\bottomrule
\end{tabular}}
\caption{\label{tab:training-stats}\textbf{Training Statistics Comparison.}  
While \method requires longer 
training time ($\sim$70 vs $\sim$50 minutes), it achieves higher 
computational utilization efficiency (18-19\% vs 17-18\% FLOPs utilization), demonstrating more effective use of available computational resources. Due to additional rollouts from the teacher model, \method has slightly 
higher FLOPs consumption}
\end{table*}

\begin{figure*}[!t]
\begin{center}
\includegraphics[width=2\columnwidth,]{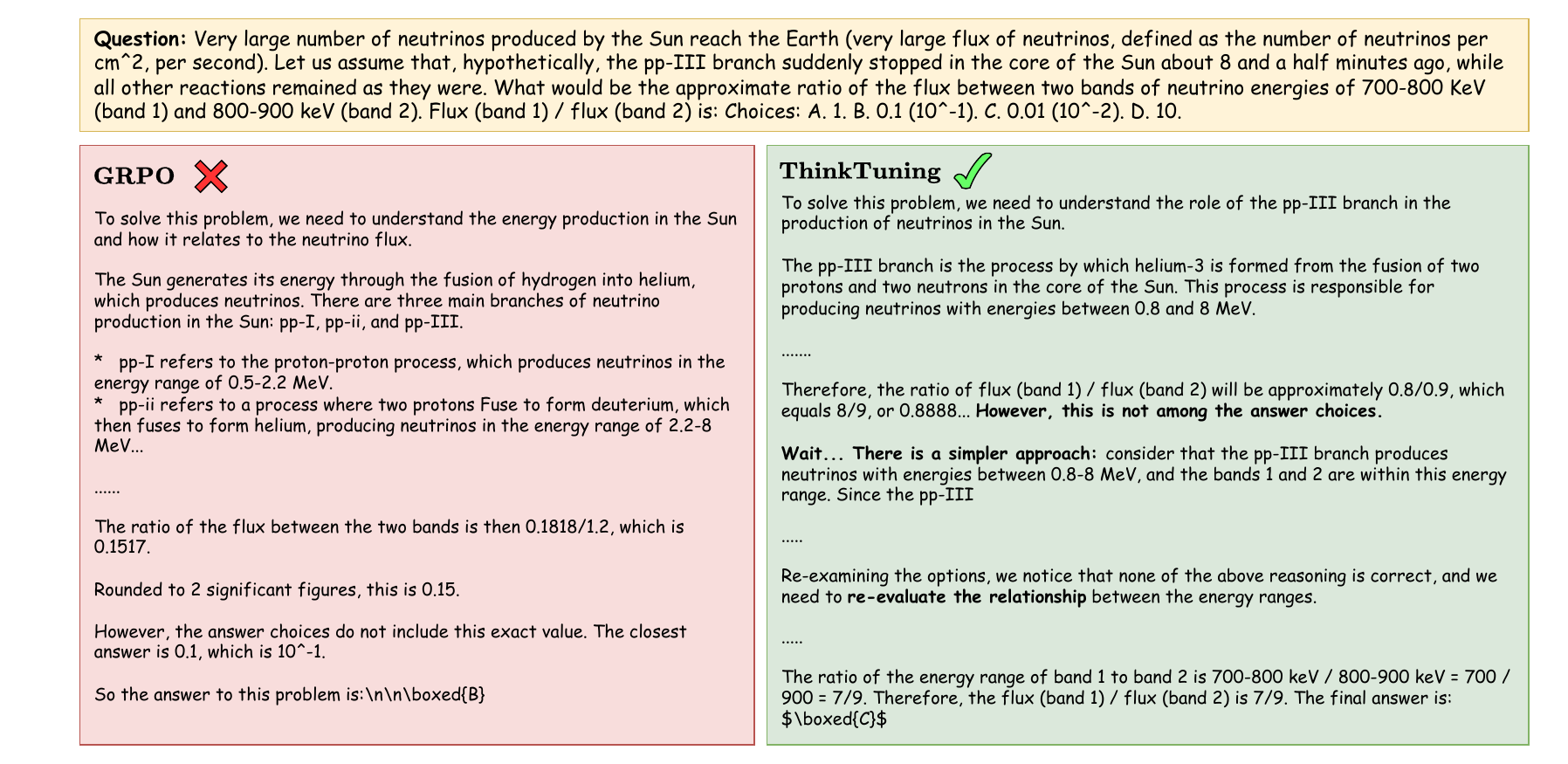}
\caption{Qualitative comparison on a GPQA-D example. The left pane shows the GRPO-trained model detecting that its computed ratio isn’t among the answer choices but then simply selecting the closest option without revisiting its reasoning, whereas the right pane illustrates \method’s self-reflective process—questioning its initial approach, re-evaluating the relationship between energy bands, and arriving at the correct flux ratio.}
\label{fig:qualitative}
\end{center}
\end{figure*}

 \method instills cognitive behaviors such as self-verification and deliberate re-evaluation, leading to notable improvements across a range of reasoning benchmarks. These reflective strategies are particularly beneficial in domains requiring multi-step scientific reasoning. Figure~\ref{fig:qualitative} illustrates a representative example from GPQA-D to highlight the qualitative impact of \method. The GRPO model recalls relevant domain knowledge but often falls short in applying it effectively to the problem at hand. In contrast, the \method-trained model shows a greater tendency to reflect on its initial reasoning, reassess intermediate steps, and adjust its approach if needed. This form of self-correction contributes to more consistent outcomes, particularly on questions that benefit from structured re-evaluation.

\end{document}